\documentclass[sigconf, nonacm]{acmart}
\renewcommand\footnotetextcopyrightpermission[1]{} 
\setcopyright{none}
\makeatletter
\renewcommand\@formatdoi[1]{\ignorespaces}
\makeatother
\AtBeginDocument{%
  }

\setcopyright{none}
\makeatletter
\renewcommand\@formatdoi[1]{\ignorespaces}
\makeatother
\acmConference[Conference acronym 'XX]{Make sure to enter the correct
  conference title from your rights confirmation email}{June 03--05,
  2018}{Woodstock, NY}
\acmISBN{978-1-4503-XXXX-X/2018/06}

\usepackage{subcaption}
\usepackage{dsfont}
\usepackage{pifont}
\usepackage[table]{xcolor}

\begin{document}

\title{Mixture of Predefined Experts: Maximizing Data Usage on Vertical Federated Learning}

\author{Jon Irureta}
\email{jirureta@ikerlan.es}
\affiliation{%
  \institution{Ikerlan Technology Research Center}
  \city{Arrasate}
  \country{Spain}
}
\affiliation{%
  \institution{University of the Basque Country UPV/EHU} 
  \country{Spain}
}

\author{Gorka Azkune}
\email{gorka.azcune@ehu.eus}
\affiliation{%
  \institution{HiTZ Basque Center for Language Technology - Ixa, University of the Basque Country UPV/EHU}
  \country{Spain}}

\author{Jon Imaz}
\email{jimaz@ikerlan.es}
\affiliation{%
  \institution{Ikerlan Technology Research Center}
  \city{Arrasate}
  \country{Spain}
}

\author{Aizea Lojo}
\email{alojo@ikerlan.es}
\affiliation{%
 \institution{Ikerlan Technology Research Center}
 \city{Arrasate}
 \country{Spain}}

\author{Javier Fernandez-Marques}
\email{javier@flower.ai}
\affiliation{%
  \institution{Flower Labs \& University of Cambridge}
  \city{Cambridge}
  \country{United Kingdom}}

\renewcommand{\shortauthors}{Irureta et al.}

\begin{abstract}
Vertical Federated Learning (VFL) has emerged as a critical paradigm for collaborative model training in privacy-sensitive domains such as finance and healthcare. However, most existing VFL frameworks rely on the idealized assumption of full sample alignment across participants, a premise that rarely holds in real-world scenarios. To bridge this gap, this work introduces Split-MoPE, a novel framework that integrates Split Learning with a specialized Mixture of Predefined Experts (MoPE) architecture. Unlike standard Mixture of Experts (MoE), where routing is learned dynamically, MoPE uses predefined experts to process specific data alignments, effectively maximizing data usage during both training and inference without requiring full sample overlap. By leveraging pretrained encoders for target data domains, Split-MoPE achieves state-of-the-art performance in a single communication round, significantly reducing the communication footprint compared to multi-round end-to-end training. Furthermore, unlike existing proposals that address sample misalignment, this novel architecture provides inherent robustness against malicious or noisy participants and offers per-sample interpretability by quantifying each collaborator's contribution to each prediction. Extensive evaluations on vision (CIFAR-10/100) and tabular (Breast Cancer Wisconsin) datasets demonstrate that Split-MoPE consistently outperforms state-of-the-art systems such as LASER and Vertical SplitNN, particularly in challenging scenarios with high data missingness.
\end{abstract}



\received{20 February 2007}
\received[revised]{12 March 2009}
\received[accepted]{5 June 2009}

\maketitle

\section{Introduction}

\begin{figure}[]
    \centering
    \begin{subfigure}{0.48\columnwidth}
        \centering
        \includegraphics[width=\textwidth]{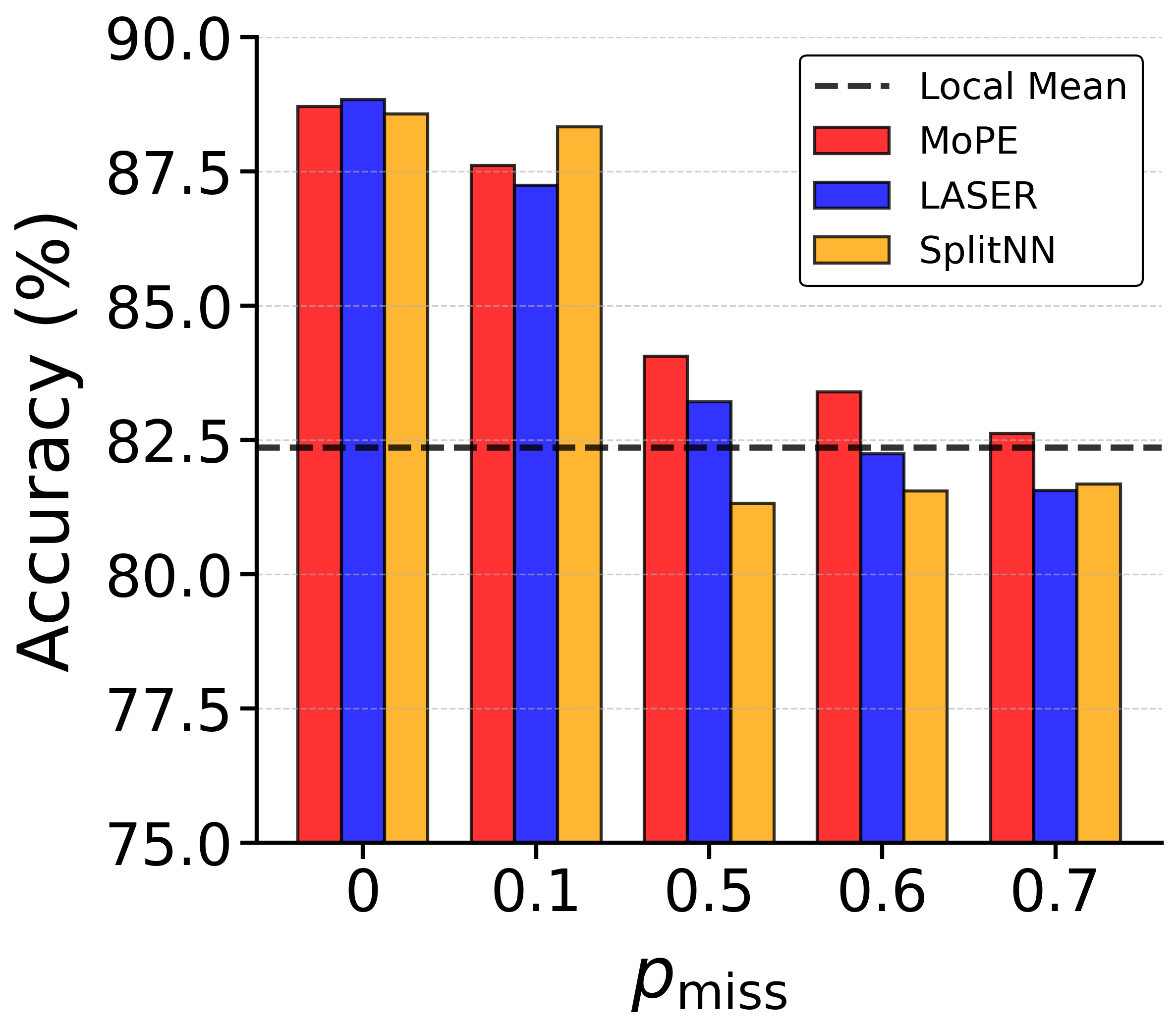}
        \label{subfig: summary accuracy}
    \end{subfigure}
    \hfill
    \begin{subfigure}{0.48\columnwidth}
        \centering
        \includegraphics[width=\textwidth]{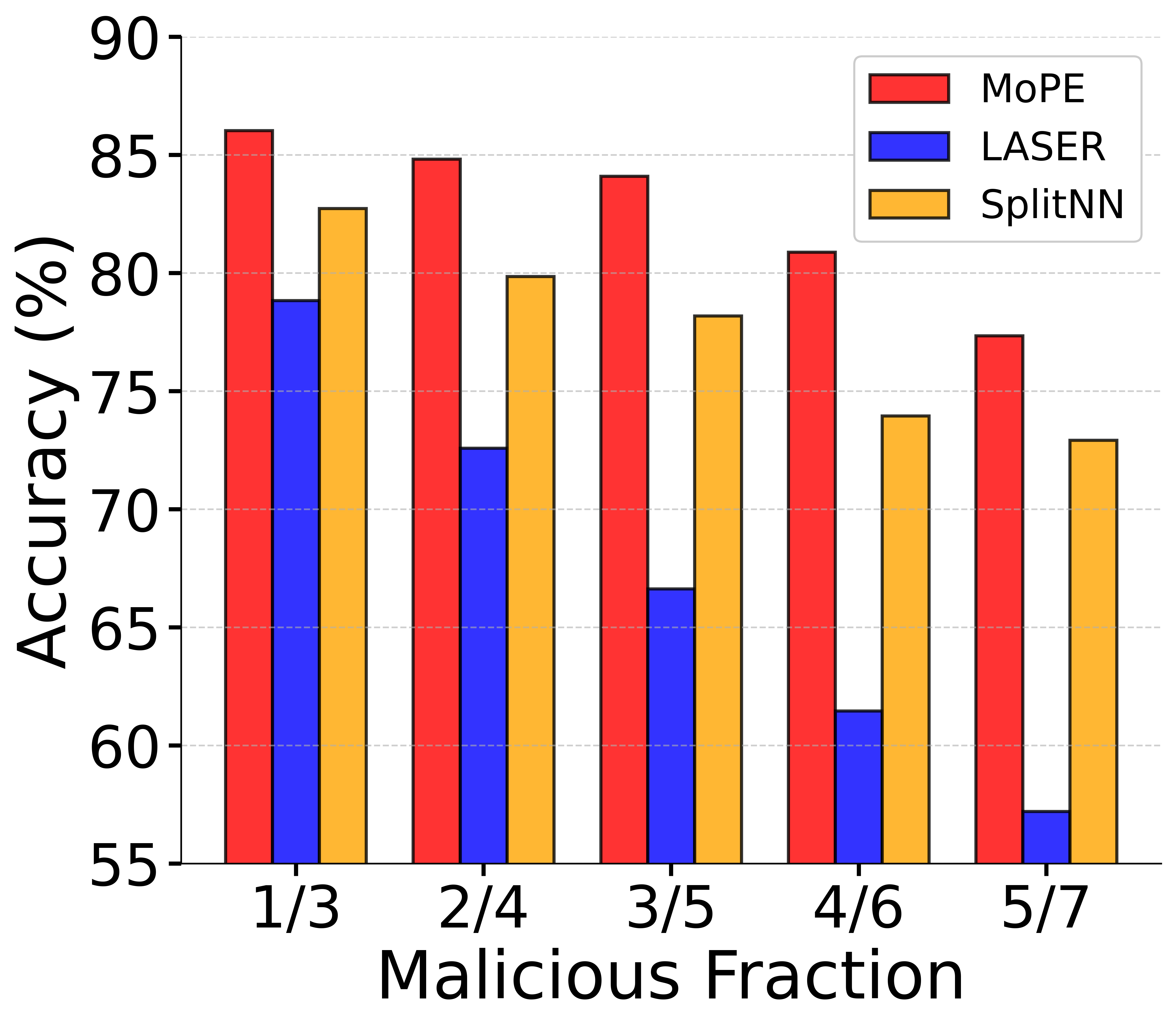}
        \label{subfig: summary decay}
    \end{subfigure}
    \vspace{-5mm}
    \caption{Mean performance of different VFL methods across CIFAR-10 and CIFAR-100. The \textit{left} plot shows mean accuracies under varying data missingness ratios, where a higher \(p_{miss}\) implies more missing data, while the \textit{right} plot shows mean accuracies as the fraction of noisy or malicious participants in the federation increases.}
\label{fig:summary plot}
\end{figure}

Federated Learning (FL) aims to enable collaborative model training without requiring the sharing of raw data among collaborators, participants, or members of the federation \citep{kairouz2021advances}. Due to its ability to maintain data local and, therefore, private, FL is particularly relevant in domains where data privacy is not optional but mandatory, such as finance and medicine \citep{wen2023survey}.

Within FL, two primary scenarios can be distinguished based on how data is partitioned among the participants: Horizontal FL (HFL) and Vertical FL (VFL). HFL has been widely studied in the literature \citep{li2020federated, mcmahan2017communication, reddi2020adaptive},  and is applicable to cases where collaborators have a common feature space but different sample spaces \cite{mcmahan2017communication}. In contrast, VFL, which has received less attention, is applicable when the dataset is vertically partitioned, that is, participants share the same sample space but possess different features \cite{khan2025vertical}. Beyond differing in data partitioning, these two scenarios also vary in their training approaches. In HFL, a global model is typically trained and shared among all participants. In VFL, however, the model is usually trained for a single participant, referred to as the active participant, who has access to the labels. The other participants, known as passive participants, contribute their features to support the training process \cite{liu2024vertical}. 

Despite its potential, most VFL proposals suffer from a critical sample alignment bottleneck. Most existing frameworks operate under the idealistic assumption of perfect sample overlap across participants. In real-world deployments, such as cross-hospital collaborations, the intersection of patients across institutions is often sparse, making standard VFL methods inefficient or inapplicable. Current state-of-the-art approaches \citep{valdeira2024vertical, huang2023vertical} do not make this assumption but still require frequent information exchanges between participants, leading to significant communication overhead and operational complexity \citep{irureta2024towards}. Furthermore, as the number of passive participants grows, the system becomes vulnerable to malicious or noisy data contributions, with little or no insight into which participants are making the model's decisions \cite{cui2024survey}.

To address these challenges, we propose Split-MoPE, a novel VFL framework that combines Split Learning \citep{vepakomma2018split} with a newly introduced Mixture of Predefined Experts (MoPE), an adapted version of the Mixture of Experts (MoE). Unlike standard MoE, where routing is learned dynamically, MoPE uses a set of experts specifically assigned to process different data alignments that naturally arise in VFL scenarios. By leveraging pretrained local encoders, Split-MoPE allows for training with a single information exchange between participants, reducing overall system complexity and communication footprint. Furthermore, Split-MoPE outperforms state-of-the-art VFL frameworks as data missingness increases, while offering superior robustness to malicious or noisy participants (Figure \ref{fig:summary plot}).

Our main contributions are the following:
\begin{itemize}

\item \textbf{Alignment-agnostic VFL:} We propose a novel VFL framework that does not require full sample alignment. Our proposal is on par with well-established state-of-the-art VFL methods such as LASER \citep{valdeira2024vertical} and Vertical SplitNN\footnote{For brevity, we will use SplitNN for the rest of the paper.} \citep{vepakomma2018split} when data alignment is perfect, but significantly outperforms them as data missingness increases, with differences up to 5 percentual points depending on the dataset.
    
\item \textbf{Robustness and accountability:} We empirically show that MoPE's gating mechanism inherently limits the influence of malicious or non-informative participants, reducing their impact on model performance by up to $25\times$ compared with LASER and SplitNN. Furthermore, the architecture provides per-sample interpretability by quantifying each participant's contribution.

\item \textbf{Modularity and Decoupled Training:} We introduce a modular architecture that leverages pretrained modality-specific encoders to facilitate decoupled training. This design eliminates the need for end-to-end training across parties, enabling training with a single communication round without sacrificing the integrity of local representations or the richness of the extracted features.
    
\end{itemize}

\section{Background and Related Work} \label{sec: background}
In this section, we lay the theoretical foundations for understanding the proposed Split-MoPE framework. We begin by defining the mechanics of VFL and its challenges regarding data alignment and communication efficiency in existing literature. We then review the MoE architecture, which serves as the basis for our proposed method for handling different data partitions in VFL.
\subsection{Vertical Federated Learning}
\begin{figure}[h]
    \centering
    \begin{subfigure}{0.48\columnwidth}
        \centering
        \includegraphics[width=\textwidth]{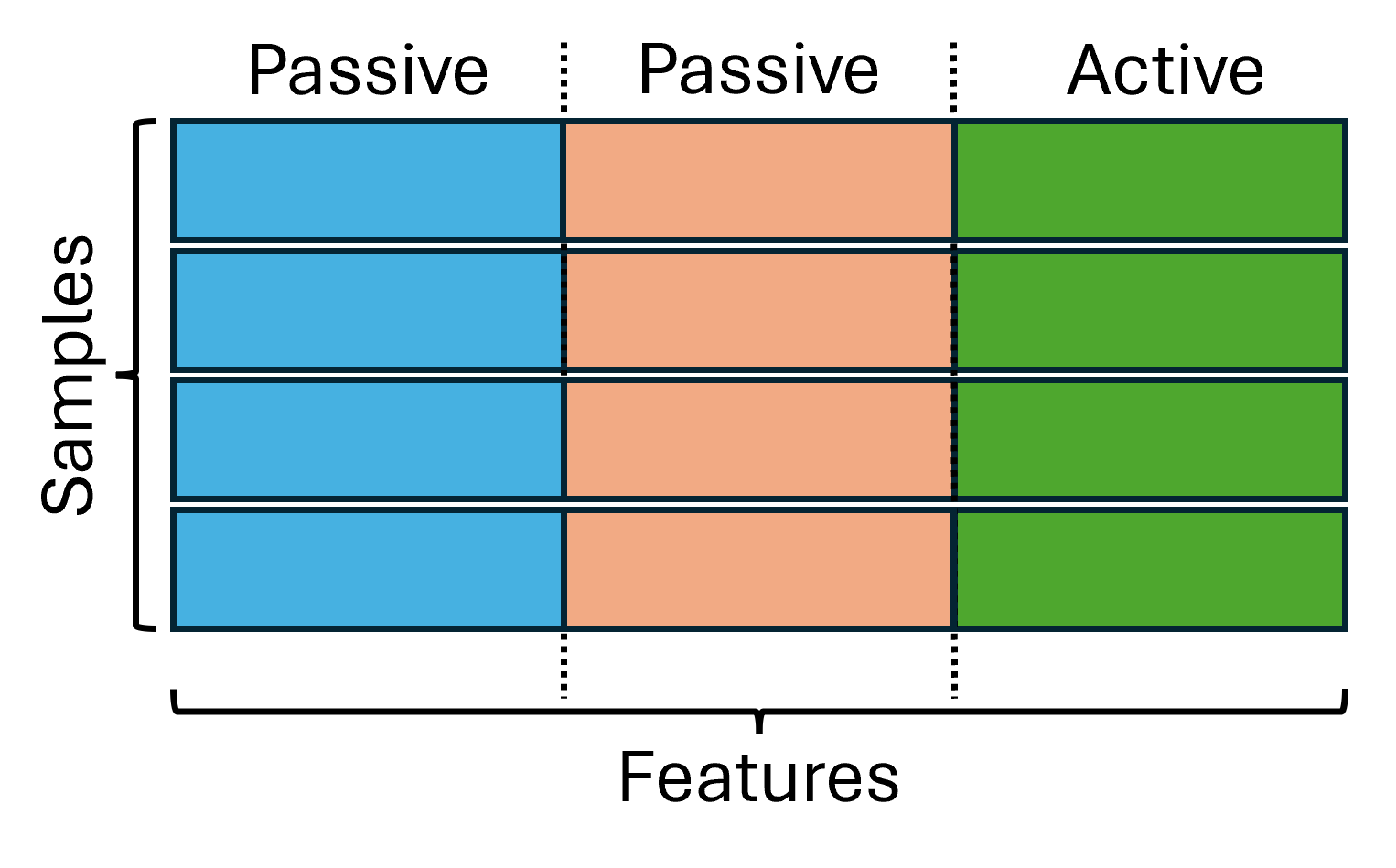}
        \caption{Classical VFL data partition.}
        \label{subfig: classical vfl partition}
    \end{subfigure}
    \hfill
    \begin{subfigure}{0.48\columnwidth}
        \centering
        \includegraphics[width=\textwidth]{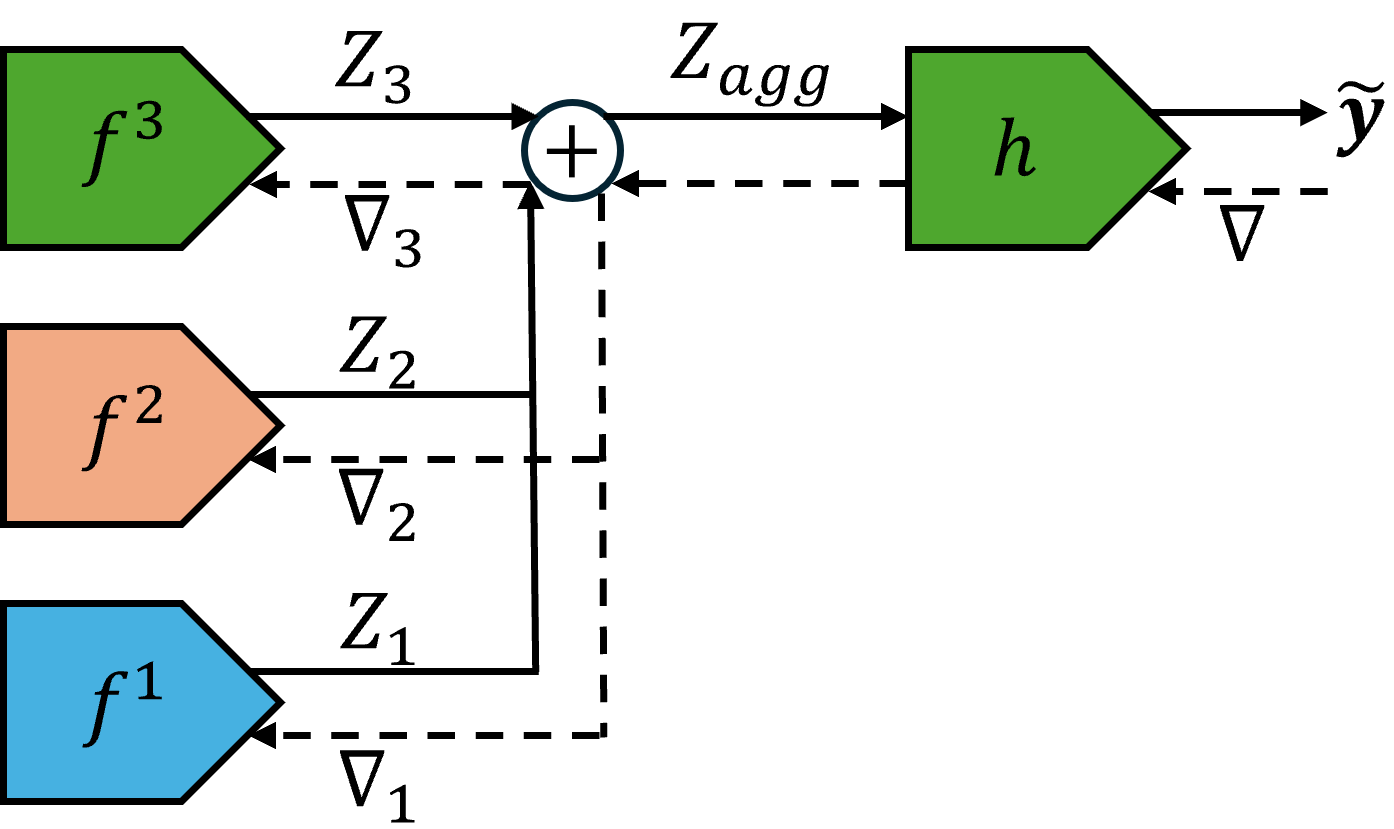}
        \caption{SplitNN.}
        \label{subfig: vertical splitnn}
    \end{subfigure}
    
    \caption{(a) Classical VFL data partition, where the sample space is identical for every participant. (b) SplitNN's high-level architecture with the forward pass (solid line) and the backpropagation steps (dashed line).}
    \label{fig:splitnn-classical-partition}
\end{figure}

Classical VFL is defined by the condition that all participants share the same sample space, as depicted in Figure \ref{subfig: classical vfl partition}. Within this framework, split learning, specifically the SplitNN approach, is widely adopted \citep{vepakomma2018split, ceballos2020splitnn}. To implement neural networks for the VFL setting using split learning, each participant maintains a local encoder or feature extractor, $f^k$. Additionally, the active participant is responsible for defining a classification head, $h$, that produces predictions.

Consider an example setup with three participants, as illustrated in Figure \ref{subfig: vertical splitnn}. For the forward pass (represented with solid lines), each participant processes its local data batch $X_k$ to produce embeddings $Z_k$ via its feature extractors: $Z_k=f^k(X_k)$. These embeddings are then transmitted to the active participant, who aggregates them into a single representation, $Z_{agg}$, ideally through concatenation to maximize performance \citep{li2023fedvs}. The active participant then applies the classification head: $\tilde{\mathbf{y}}=h(Z_{agg})$, generating the final prediction.

Using the available labels, the active participant computes the error and its gradient with respect to the model parameters, updating the classification head ($h$) accordingly. To complete the update process, the active participant also calculates the gradients relevant to each feature extractor ($f^k$), and communicates these gradients to the corresponding passive participants, allowing them to update their local models (showed using dashed lines on Figure \ref{subfig: vertical splitnn}).

\begin{figure}[h]
  \centering
  \includegraphics[width=0.50\linewidth]{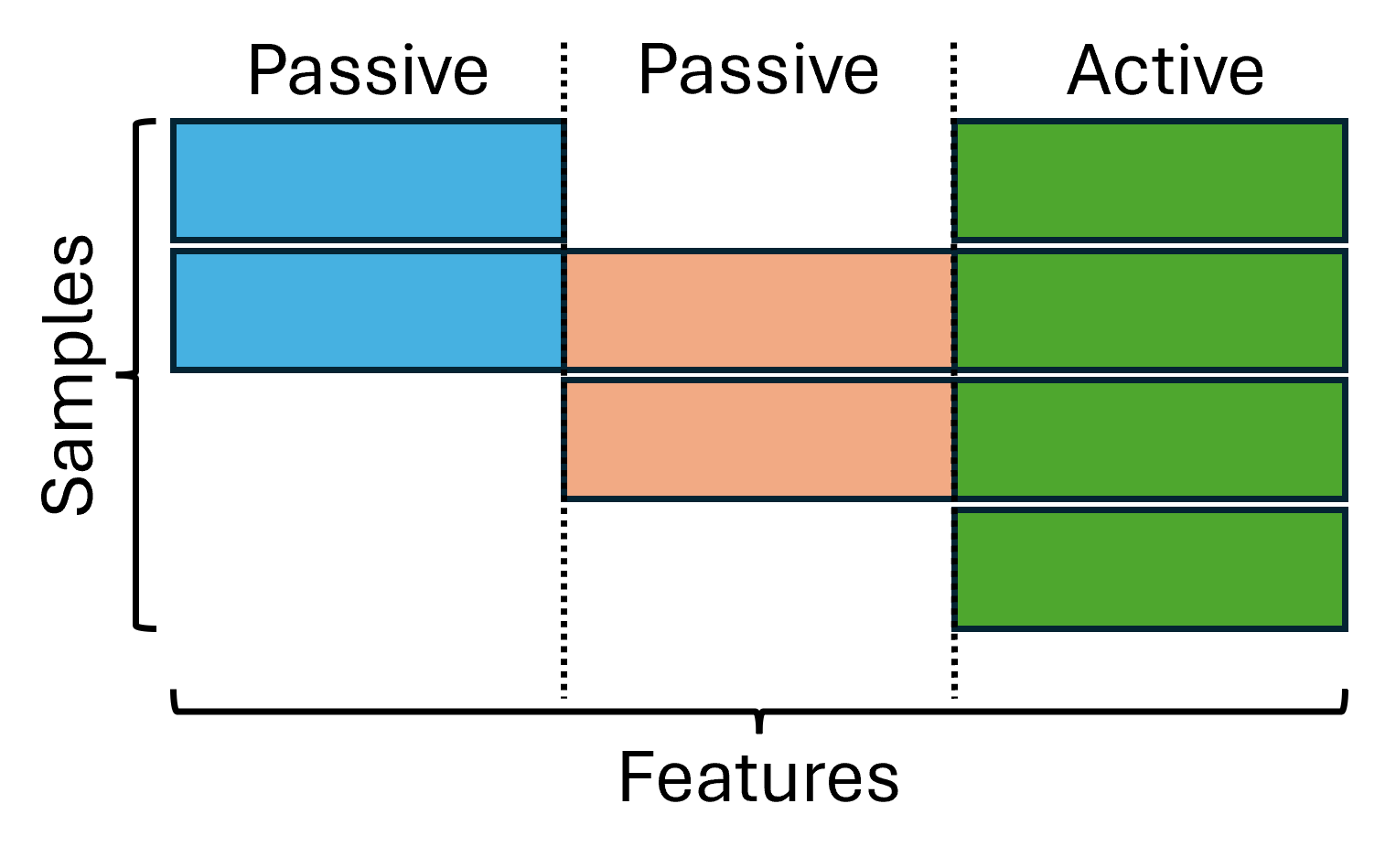}
  \caption{Real data partition of VFL where samples among participants are not fully aligned.}
  \label{fig: real vfl partition}
\end{figure}

When there are missing feature blocks as represented in Figure \ref{fig: real vfl partition}, approaches such as SplitNN or alternatives that have been built on top of the assumption of full sample alignment, suffer great performance drop \citep{valdeira2024vertical}.

More recent proposals that focus on the algorithmic aspect of VFL do not fall on the assumption of an ideal, fully aligned scenario, but still, training and doing inference with every sample remains relatively unexplored. FedHSSL \citep{he2024hybrid} and FedCVT \citep{kang2022fedcvt} aim to improve the effectiveness of the process by incorporating the knowledge of unaligned samples to the training using self-supervised learning, but still fail to use non-aligned samples for inference, which is critical for VFL's applicability in real scenarios. Furthermore, these approaches require frequent information exchange and synchronization between participants, which further increases the complexity to apply them on non-simulated environments.

Reducing the number of required communication rounds of the aforementioned proposals, \citet{sun2023communication} propose One-shot/Few-shot VFL, which limits the number of information exchanges between participants and thus, synchronization, but still fail to do inference on non-aligned samples. While maintaining limited information exchanges, VFedTrans \citep{huang2023vertical} and APC-VFL \citep{irureta2024towards} allow to use unaligned samples not only for training, but inference as well. Even if these works take an important step towards making VFL applicable in real scenarios, their extensions to scenarios with more than two participants are non-trivial, as every data partition that arises on those cases is not taken into account, and include extra hyperparameter w.r.t the traditional split learning based approach. This increases the overall complexity of the proposal.

Currently, in terms of performance, the state-of-the-art is LASER-VFL \cite{valdeira2024vertical}, which incorporates a sampling-based aggregation mechanism into the traditional SplitNN. Even if effective, hyperparameter free\footnote {An approach is denoted as hyperparameter free if it requires no extra hyperparameters with respect to Vertical SplitNN.}, and applicable on federations of any size, this work requires frequent information exchanges between collaborators and does not respect the inherent data partitions of a VFL set-up, neither for training nor inference, which could harm the overall performance.

With Split-MoPE we have introduced an effective, hyperparameter-free proposal that is efficient in terms of communication overhead and respects the multiple data alignments of VFL scenarios, maximizing performance. For this task, we combine SplitNN with a variant of the Mixture of Experts architecture \citep{jacobs1991adaptive, jordan1994hierarchical}.

\subsection{Mixture of Experts}
The Mixture of Experts architecture aims to reduce the number of active parameters in a model by dividing a neural network into sub-networks, known as \textit{experts}. This results in an architecture that allows efficient scaling of model parameters without increasing computational demand \citep{cai2025survey}. Because of that, MoE-based architectures are widely used on LLMs \citep{jiang2024mixtral, dai2024deepseekmoe} or CV applications \citep{riquelme2021scaling, lou2021sparse}.


A MoE architecture is composed of a set of $N$ experts $\mathcal{E} = \{E_1, E_2, ..., E_N\}$, and a router $G$. The number of experts used during the forward pass is generally a fixed number $M$, where usually $M\ll N$ \cite{jin2024moe++}. Taking this into account, the forward pass can be formalized with Equation \ref{equ: moe forward}:

\begin{equation}
    \label{equ: moe forward}
    \mathbf{z} = \sum_{i=1}^N g_i\cdot E_i(\mathbf{x})
\end{equation}
where,
\begin{description}
  \item[$
    g_i = 
    \begin{cases}
      \mathrm{Softmax}(G(x))_i, & \textnormal{if } G(x)_i \in \mathrm{Top}\text{-}M(\{G(x)_i | 1 \leq i \leq N\}) \\
      0, & \textnormal{otherwise.}
    \end{cases}
  $]
\end{description}

Note that for consistency and to avoid notation redundancies, we denote that \textit{top-M} experts are used, but frequently, $K$ is used instead of $M$. Based on the number of experts used during the forward pass, $M$, different routing strategies exist, being the most common ones: Top-1 (only the expert to which most weight has given the router is activated) \citep{fedus2022switch}, Top-2 \citep{lepikhin2020gshard} or Top-p, where $M$ is computed based on the amount of experts needed to reach a minimum total weight, i.e., the sum of $g_i$s \citep{huang2024harder}. Even if selecting subsets of experts is the most common choice, it must be noted that using all experts is also possible, which is known as \textit{Dense Mixture of Experts} \citep{nie2021dense}.

These routing strategies shape the behavior of the MoE architecture as well as the way the experts are trained, which makes it a critical decision to ensure the correct functioning of the layer \citep{dikkala2023benefits}.

\section{Problem Definition}
Consider a VFL scenario with $K$ participants, where $\mathcal{K}:=[K]$ is the set of collaborators that intend to train a classification model using a vertically partitioned dataset. Being $\mathcal{X}_k$ the feature space, $\mathcal{Y}_k$ the label space, and $\mathcal{I}_k$ the ID space for the $k$-th participant, the distributed datasets can be defined as follows: $\mathcal{D}_k = \{\mathbf{i}_k, X_k\}\forall k\ne K$, whereas, $\mathcal{D}_K = \{\mathbf{i}_K, X_k, \mathbf{y}_K\}$, where $\mathbf{i}_k\in\mathcal{I}_k,\;X_k\in\mathcal{X}_k,\; \mathbf{y}_k\in\mathcal{Y}_k$. Note that we set the participant $k = K$ as the active participant, i.e., the one with access to the labels and willing to make predictions. Given that the dataset is vertically partitioned, the features each participant has access to differ, but the ID space is common, i.e. $\mathcal{X}_i\ne\mathcal{X}_j$, $\mathcal{I}_i=\mathcal{I}_j \forall i\ne j$.

Taking into account that the ID space is common but the IDs themselves do not need to be identical, $\mathbf{i}_i\ne\mathbf{i}_j$, samples can either be \textit{aligned} or \textit{unaligned}, which we assume is known by the participants after conducting a private set intersection (PSI) process based on the IDs \citep{morales2023private}. Note that this is a common assumption on those proposals that focus on the algorithmic aspect of VFL \citep{he2024hybrid, kang2022fedcvt, sun2023communication}. Since a sample can be present on every participant, a subset of them, or in a single one, all such possible combinations are defined by the powerset, $\mathcal{P}(\cdot)$, of the set of participants, i.e., $\mathcal{P}(\mathcal{K})$.  

It must be noted that not every combination is \textit{interesting}, since a sample which is unknown to the active participant will never be used during inference. Because of that, we define $\tilde{\mathcal{P}}(\mathcal{K})$, the set of every \textit{interesting combination}:

\begin{equation}
    \label{equ: VFL alignments}
    \tilde{\mathcal{P}}(\mathcal{K}) := \{\mathcal{S}\in\mathcal{P}(\mathcal{K})|K\in\mathcal{S}\}
\end{equation}

$\tilde{\mathcal{P}}(\mathcal{K})$ contemplates all the \textit{aligned} samples that are present on the dataset of the active and at least one passive participant, as well as those \textit{unaligned} samples that are present on the dataset of the active participant exclusively.

In this work, we aim to maximize model performance by respecting the data alignments, i.e., different combinations, that exist in real VFL scenarios (see Equation \ref{equ: VFL alignments}), where datasets are not necessarily fully \textit{aligned} between participants, scenarios, where not every ID is present on the dataset of every participant, while making the process efficient from the communication footprint standpoint. 

\section{Method}
\label{sec:method}
\begin{figure}[h]
  \centering
  \includegraphics[width=\linewidth]{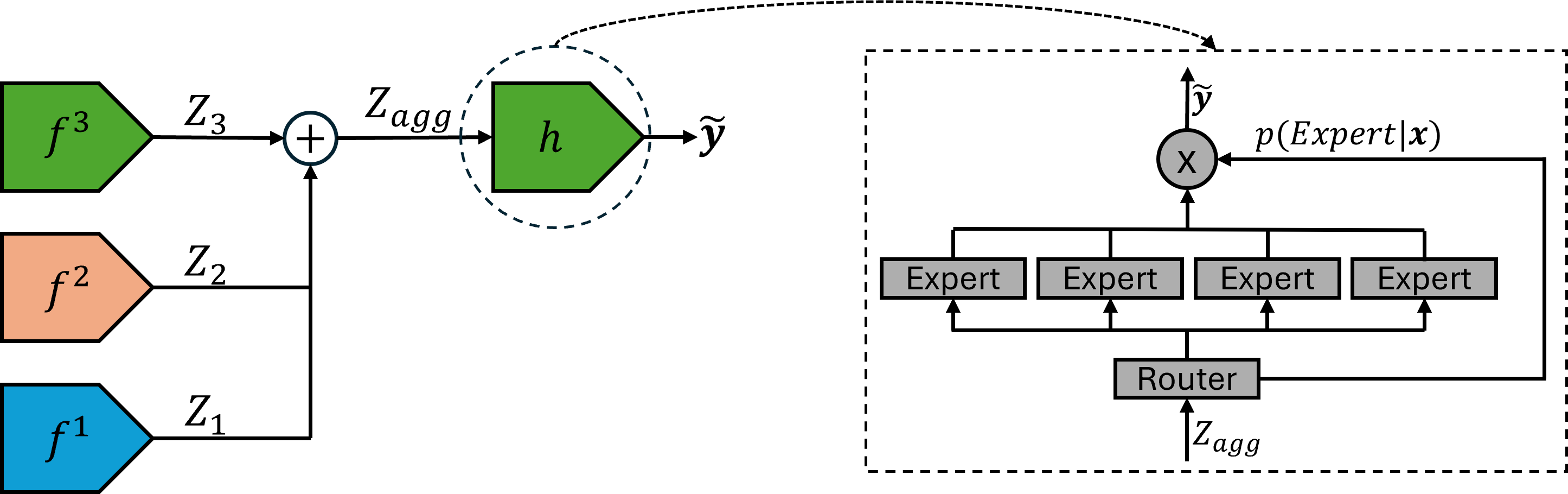}
  \caption{Our proposed Split-MoPE, combining Split learning with a modified classification head, $h$, which consists of a Mixture of Predefined Experts (MoPE) layer.}
  \label{fig: split mope}
\end{figure}

Our proposal, called Split-MoPE, is represented in Figure \ref{fig: split mope}. Split-MoPE combines the regular SplitNN with a modified Mixture of Experts layer, the Mixture of Predefined Experts (MoPE) layer, which substitutes the classical classification head.

\subsection{Desired Properties of VFL Systems}

\label{subsec:desing-principles}
To address the challenges of VFL, specifically data misalignment and communication overhead, the design of Split-MoPE is guided by the following architectural desiderata:

\begin{itemize}
    \item \textbf{Vertical Baseline Recovery:} In a fully aligned scenario, and assuming that all participants contribute relevant information, an ideal VFL system should converge to the performance of a classifier system trained on all the aligned features and labels.
    \item \textbf{Local Robustness:} An ideal VFL system must remain resilient against noisy or missing contributions from passive participants. A robust architecture should converge to the performance of a locally trained model if the collaboration does not report benefits. 
    \item \textbf{Adaptive Alignment Weighting:} When full-sample alignment does not exist but the collaborative approach shows benefits, the system should maximize overall performance. For that end, it should take into account the informativeness of each partition defined by $\tilde{\mathcal{P}}(\mathcal{K})$.
    \item \textbf{Communication Efficiency (Single-Round Training):} To reduce the communication overhead, the system must minimize the number of required information exchanges. The architecture should require only a single transmission of representations from passive to active participants.
\end{itemize}



\subsection{Split-MoPE}
Split-MoPE follows the coarse-level design of SplitNN \cite{ceballos2020splitnn}, modifying two crucial aspects of it to fulfill the requirements identified in Section \ref{subsec:desing-principles}: i) combining decoupled training \citep{irureta2024towards} with pretrained (and frozen) backbone encoders, to effectively address single-round training, and ii) replacing the classic classification head with our novel Mixture of Predefined Experts (MoPE) head, to address the rest of the requirements.
\smallskip

\noindent \textbf{Combining pretrained encoders and decoupled training:} To address the first requirement, communication efficiency, Split-MoPE leverages pretrained backbones as feature extractors. By employing frozen, off-the-shelf models, the need for backpropagation through the entire network is eliminated, allowing training to be completely omitted for the feature extractor layers. Consequently, the communication overhead is reduced to a single forward pass of representations, satisfying the requirement for a \textbf{single-round training} process. This approach is particularly powerful in the current landscape of deep learning, where a vast ecosystem of pretrained backbones exists across multiple modalities and scales, which allows to choose them based on the target application.

\smallskip

\begin{figure}[h]
  \centering
  \includegraphics[width=0.7\linewidth]{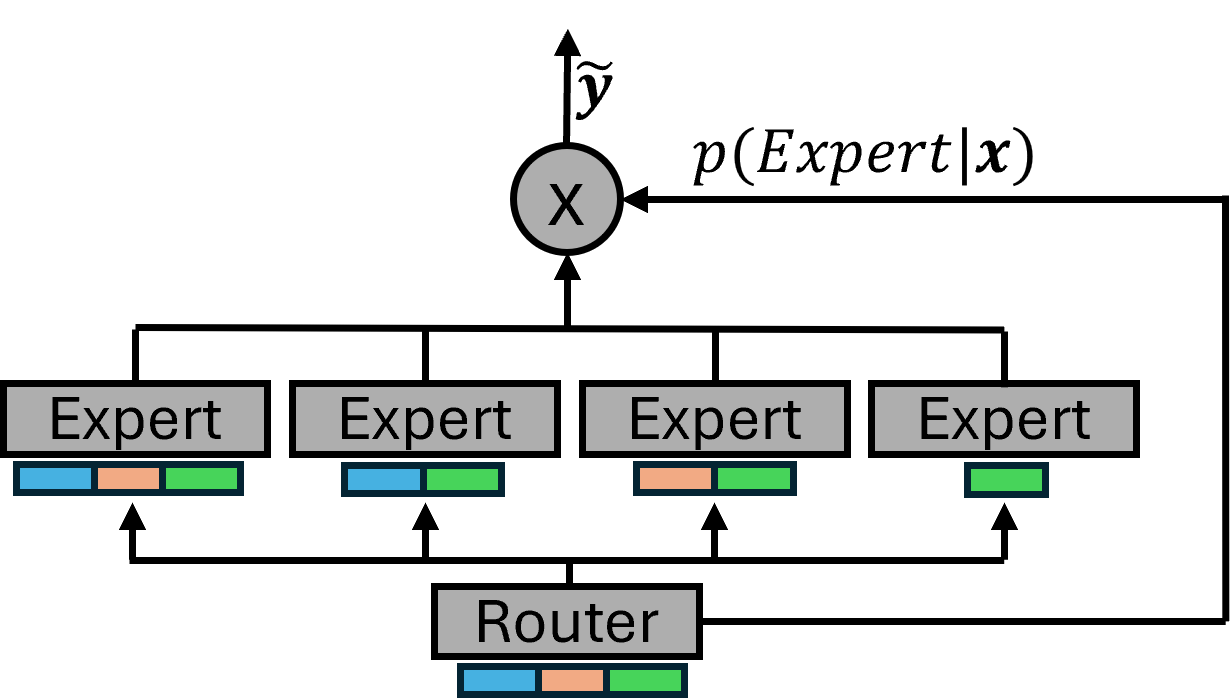}
  \caption{Representation of the forward pass in MoPE. The embeddings corresponding to each participant's information are represented in different colors (active in green), and each Expert in the layer processes a combination of them.}
  \label{fig: mope forward fully aligned}
\end{figure}

\noindent \textbf{The proposed MoPE head:} Unlike traditional MoE, where experts learn to specialize in latent data patterns or tokens during training, MoPE introduces structural constraints tied directly to the data partitions of VFL. In MoPE, the domain of each expert is \textit{predefined} by a specific subset of $\tilde{\mathcal{P}}(\mathcal{K})$ (see Figure \ref{fig: mope forward fully aligned}). This results in a one-to-one mapping where the number of experts is strictly determined by the number of \textit{interesting} alignments: $|\mathcal{E}| = |\tilde{\mathcal{P}}(\mathcal{K})|$.

While the internal structure and data-routing logic diverge significantly from the MoE paradigm to solve the unique challenges of VFL, the two architectures are identical in terms of their core building blocks: a router, $G$, and a set of experts, $\mathcal{E}$. In our case, the router is instantiated by a two-layer MLP, and each expert is a shallow MLP for classification (see the details in Appendix \ref{appendix: arch and params}).



The rationale behind this approach is that the optimal embedding aggregation strategy for SplitNN is concatenation, as no information is lost in the process. This poses a challenge when datasets are not fully aligned, as the input dimension of the classification head should dynamically adapt based on the sample it processes. A straightforward alternative would be to define the input dimension of the classification head with the maximum possible dimension of $Z_{agg}$, that is, the dimension of a fully aligned sample, and apply padding when there is missing information; even if possible, this results in a great performance drop, as explained in Section \ref{sec: background}.

A natural solution to maximize performance, respecting the data partitions of VFL, would be to define $|\tilde{\mathcal{P}(K)}|$ classifiers, one per partition. Such an approach would pose two problems: training each classifier independently would result in time-consuming computations, and combining the outputs of multiple classifiers to obtain a final prediction would not be trivial.

Aiming to address those challenges, our MoPE not only trains all the classifiers, experts, at once but also learns how they interact with each other to obtain an ensemble prediction via the router. Therefore, there is no need to train each model independently, nor to learn how to aggregate their predictions, as the MoPE does so during the training process.

\smallskip





\noindent \textbf{Forward pass of MoPE:} Assuming that $\mathbf{z}_{agg}\in\mathbb{R}^N$ is the concatenation of the embeddings that correspond to a sample of every participant, i.e. the aggregated embedding of a fully aligned sample, the router decomposes it on every possible sub-vector $\mathbf{z}_{agg_i}',\; i=1, ..., |\tilde{\mathcal{P}}(\mathcal{K})|$, that correspond to the alignments defined by $\tilde{\mathcal{P}}(\mathcal{K})$. Having these sub-vectors, each one is forwarded through the $i$-th expert, which specializes on that specific alignment, as represented on Figure \ref{fig: mope forward fully aligned}.

In those cases where $\mathbf{z}_{agg}$ has missing information, i.e., it is not a fully aligned sample, we use padding and apply the same methodology as the one for the fully aligned samples. Instead of discarding experts without \textit{full information}, by applying padding, we allow the entire layer to be trained equally, maximizing interactions between experts and ensemble predictions.

Taking this into account, we can rewrite the generic MoE's forward pass from Equation \ref{equ: moe forward} as follows:
\begin{equation}
    \tilde{\mathbf{y}} = \sum_{i=1}^{|\tilde{\mathcal{P}}(K)|}g_i\cdot E_i(\mathbf{z}'_{agg_i})
\end{equation}

Unlike standard MoE architectures, where gating weights, $g_i$, are computed via Softmax (see Equation \ref{equ: moe forward}), our approach uses Sigmoid activation on the router output. This shift allows gating weights to be determined independently, encouraging experts to collaborate rather than engage in mutually exclusive selection \citep{csordas-etal-2023-approximating}.

Since the Sigmoid activation is used for the router: $0 \le g_i \le 1 \ \forall i$ but $\sum_i g_i \neq 1$. Therefore, we normalize the weighted expert outputs by dividing them by the total mass, i.e., $\sum_{i=1}^{|\mathcal{E}|}g_i$. Applying the normalization, we ensure that $\sum_y p(y|\mathbf{z}_{agg}) = 1$.
\smallskip

\noindent \textbf{Training dynamics of MoPE:} MoPE's training is carried out without constraints, in a self-regulated manner. This differs significantly from the training approach of MoEs, which use additional loss terms to avoid phenomena such as routing collapse \citep{shazeer2017outrageouslylargeneuralnetworks, chi2022representation}. 

These loss terms not only introduce additional hyperparameters that need to be tuned but also induce certain behavior in a layer, which is detrimental to our MoPE, given the unique characteristics of a VFL setting. Having a router that learns \textit{freely}, it is permitted to become highly sparse or highly collaborative as the data requires, directly enabling both \textbf{Vertical Baseline Recovery}  (by selecting only the global expert) and \textbf{Local Robustness} (by selecting only the local expert). Furthermore, this freedom allows for \textbf{Adaptive Alignment Weighting}, as the router can dynamically ensemble multiple experts in intermediate scenarios to maximize performance based on the informativeness of each partition. 

Finally, this self-regulated weighting mechanism introduces inherent interpretability to the system. By inspecting the learned weights, the contribution of each participant becomes transparent, addressing one of the major problems in VFL \citep{cui2024survey}.

\section{Experimental Set-Up}\label{sec: experimental set-up}
Following the best practices of the VFL domain \citep{valdeira2024vertical}, we detail the evaluation protocol we use to evaluate our proposal in this section.


\subsection{Datasets and Data Partitions} \label{subsec:data and partitions}
For the experimental process, we have chosen three datasets across two modalities (vision and tabular) that are widely used to evaluate VFL proposals, aiming to showcase the adaptability of our approach across diverse use cases. CIFAR-10 and CIFAR-100 \citep{krizhevsky2009learning} are the selected vision datasets, whereas Breast Cancer Wisconsin \cite{breast_cancer_wisconsin_(diagnostic)_17} is the one of tabular domain. We would like to note that, although tabular data seems to fit best with vertically partitioned datasets, image classification datasets are widely used to evaluate VFL proposals, as they are arguably more challenging \citep{sun2023communication}.

Since these datasets are not VFL-native, they need to be split. For tabular datasets, this split is natural, as each participant is assigned to a different set of columns. On the other hand, for the vision datasets, we split the images into patches, assigning each patch to a different participant in the federation as done in LASER \citep{valdeira2024vertical}.

To simulate incomplete sample alignment, we employ a missing completely at random (MCAR) mechanism, as proposed by \citet{valdeira2024vertical}. Specifically, we introduce a parameter, $p_{miss}$, that controls the probability that a passive participant cannot access its data portion, thereby rendering the sample unaligned. When the missingness probability is set to 0, the scenario corresponds to perfect sample alignment. Conversely, setting $p_{miss}$ to 1 models a scenario in which only the active participant observes the sample.

Given that in the VFL setting, unlike in the HFL, federations are small, we define scenarios with two participants and split the datasets accordingly, as is commonly done in the area \citep{sun2023communication, huang2023vertical, irureta2024towards}.

\subsection{Leveraging Pretrained Feature Extractors}
As stated previously (Section \ref{sec:method}), to reduce communication overhead while maintaining representation quality, pretrained models have been employed. For the CIFAR datasets, DINOv2 encoders \citep{oquab2023dinov2} have been used as backbones (we use ViT-S for all experiments, unless stated otherwise), and for the tabular dataset, we use the Qwen3-Embedding model \citep{zhang2025qwen3} (the 4B model is used by default). While using pretrained vision models as feature extractors is common practice, we apply a parallel logic to tabular data by transforming structured rows into semantic strings, thereby enabling the use of high-capacity text embedding models as feature extractors. 


The transformation process is the following: Assume a tabular dataset with $M$ feature columns and $N$ samples. For each sample, we convert the row to a text with the following format: ``\texttt{Feature\_name 1: val\_1 |\ldots | Feature\_name M: val\_M}''.

Having $N$ sentences that correspond to the $N$ data rows of the tabular dataset, the embedding model is used to obtain the embedding that corresponds to each sentence, i.e., each sample. For more information about this approach, refer to Appendix \ref{appendix: tabular dataset}

\subsection{Baselines}
We compare the proposed Split-MoPE method with two alternative approaches: SplitNN \citep{vepakomma2018split} and LASER \citep{valdeira2024vertical}. The rationale for this comparison is that the former represents a well-established method for addressing the vertical federated learning problem, while the latter constitutes the current state of the art for scenarios involving vertically partitioned and partially aligned datasets. We also employ two custom versions of SplitNN and LASER (denoted as SplitNN* and LASER*), where both systems are adapted to our decoupled training process with pretrained backbone encoders. This is done for two reasons: i) we can measure the contribution of the proposed training scheme and the usage of pretrained encoders, and ii) we can compare Split-MoPE with other VFL systems on equal basis.

In addition, since the primary motivation for collaboration is to obtain a global model that outperforms a model trained locally, we also train a standalone local model. This baseline serves as a lower bound that the federated method must surpass to be considered an effective and worthwhile approach.

\section{Results}
\label{sec:results}

We will next present the main results of our experiments. All the experiments involve training Split-MoPE and the reference baselines. See full details of these training processes in Appendix \ref{appendix: arch and params}.
\smallskip

\noindent \textbf{Decoupled training with pretrained and frozen encoders is beneficial for current VFL systems:} In Table \ref{tab: end to end vs frozen} we compare the performance of SplitNN and LASER with pretrained encoders under two settings: i) applying their original end-to-end training approach where pretrained encoders are also updated during training incurring in high communication costs, and ii) keeping the encoders frozen, allowing for efficient single-turn communication training (our proposed approach marked with an asterisk). The results obtained in CIFAR-100 clearly show that end-to-end training is detrimental for performance. Decoupled training with frozen encoders achieves higher accuracies by large margins across all data alignment scenarios; refer to Appendix \ref{appendix: cifar10 frozen} for additional results. Note that the results we obtain with SplitNN* and LASER* are even better than the results reported in the reference work \citep{valdeira2024vertical} (see Appendix \ref{appendix: laser improvement}). These results not only validate our combination of pretrained and frozen encoders with decoupled training but also show that our approach can be applied to different VFL systems. For the rest of the experiments, we will always use LASER* and SplitNN* as references, given their improved performance. 

\begin{table}[h]
\centering
\caption{Test accuracy (\%) on CIFAR-100 performing end-to-end training and employing frozen backbones (*). Mean $\pm$ std. dev. over three runs is reported. Best results in bold.}
\small
\label{tab: end to end vs frozen}
\begin{tabular}{lccc} 
\toprule
 & \multicolumn{3}{c}{\textbf{CIFAR-100}} \\ 
\midrule 
 & $p_{\text{miss}}=0$ & $p_{\text{miss}}=0.1$ & $p_{\text{miss}}=0.5$ \\
\midrule 
LASER & $59.98 \pm \scriptstyle 4.62$ & $58.44 \pm \scriptstyle 3.16$ & $38.58 \pm \scriptstyle 1.93$ \\
LASER* & $\textbf{82.26} \pm \scriptstyle 0.06$ & $\textbf{79.83} \pm \scriptstyle 0.50$ & $\textbf{74.52} \pm \scriptstyle 0.54$ \\
\midrule
SplitNN & $47.55 \pm \scriptstyle 1.49$ & $46.52 \pm \scriptstyle 0.94$ & $44.33 \pm \scriptstyle 1.23$ \\
SplitNN* & $\textbf{81.53} \pm \scriptstyle 0.34$ & $\textbf{81.12} \pm \scriptstyle 0.30$ & $\textbf{72.06} \pm \scriptstyle 0.28$ \\
\bottomrule
\end{tabular}
\end{table}

\smallskip


\noindent \textbf{Split-MoPE is the new state-of-the-art to handle non-aligned data in VFL:} Table \ref{tab: results non noisy} shows the performance of different VFL approaches for increasing probabilities of data missingness for the three evaluation datasets. We report the mean accuracy (F1 for Breast Cancer Wisconsin) and the standard deviation of three runs. The trends are consistent for the three datasets, showing that Split-MoPE performs on par with the reference VFL approaches when data alignment is perfect, and degrades less than the others for higher probabilities of data missingness. SplitNN is especially sensitive to data alignment, whereas LASER shows a more robust performance, but scores consistently lower than our approach.\smallskip 

\begin{table}[h]
\centering
\caption{We report the test accuracy (\%) on CIFAR-10, CIFAR-100, and F1 score ($\times100$) on Breast Cancer Wisconsin under different missing data probabilities $p_{\text{miss}}$. Our approach is shaded in {\setlength{\fboxsep}{0pt}\colorbox{gray!20}{gray}}, values that fall below the local one are shaded in {\setlength{\fboxsep}{0pt}\colorbox{red!15}{red}}, and the best collaborative approach in \textbf{bold}.}
\footnotesize
\label{tab: results non noisy}
\begin{tabular}{lccccc}
\toprule
 & \multicolumn{5}{c}{\textbf{CIFAR-10}} \\
\cmidrule(lr){2-6}
 & $p_{\text{miss}}=0$ & $p_{\text{miss}}=0.1$ & $p_{\text{miss}}=0.5$ & $p_{\text{miss}}=0.6$ & $p_{\text{miss}}=0.7$ \\
\midrule
Local   & \multicolumn{5}{c}{$91.25 \pm \scriptstyle 0.09$} \\
\midrule
SplitNN* & $95.61 \pm \scriptstyle 0.10$ & $\textbf{95.55} \pm \scriptstyle 0.08$ & \cellcolor{red!15}$90.59 \pm \scriptstyle 0.22$ & \cellcolor{red!15}$90.55 \pm \scriptstyle 0.22$ & \cellcolor{red!15}$90.80 \pm \scriptstyle 0.13$ \\
LASER*   & $95.42 \pm \scriptstyle 0.02$ & $94.66 \pm \scriptstyle 0.42$ & $91.91 \pm \scriptstyle 0.29$ & $91.27 \pm \scriptstyle 0.29$ & \cellcolor{red!15}$90.58 \pm \scriptstyle 0.10$ \\
\rowcolor{gray!20}MoPE    & $\textbf{95.63} \pm \scriptstyle 0.13$ & $94.73 \pm \scriptstyle 0.37$ & $\textbf{92.82} \pm \scriptstyle 0.13$ & $\textbf{92.47} \pm \scriptstyle 0.25$ & $\textbf{91.98} \pm \scriptstyle 0.14$ \\
\addlinespace[3pt]
\midrule
 & \multicolumn{5}{c}{\textbf{CIFAR-100}} \\
\cmidrule(lr){2-6}
 & $p_{\text{miss}}=0$ & $p_{\text{miss}}=0.1$ & $p_{\text{miss}}=0.5$ & $p_{\text{miss}}=0.6$ & $p_{\text{miss}}=0.7$ \\
\midrule
Local   & \multicolumn{5}{c}{$73.48 \pm \scriptstyle 0.28$} \\
\midrule
SplitNN* & $81.53 \pm \scriptstyle 0.34$ & $\textbf{81.12} \pm \scriptstyle 0.30$ & \cellcolor{red!15}$72.06 \pm \scriptstyle 0.28$ & \cellcolor{red!15}$72.55 \pm \scriptstyle 0.35$ & \cellcolor{red!15}$72.56 \pm \scriptstyle 0.06$ \\
LASER*   & $\textbf{82.26} \pm \scriptstyle 0.06$ & $79.83 \pm \scriptstyle 0.50$ & $74.52 \pm \scriptstyle 0.54$ & $73.22 \pm \scriptstyle 0.42$ & $72.55 \pm \scriptstyle 0.86$ \\
\rowcolor{gray!20}MoPE    & $81.78 \pm \scriptstyle 0.35$ & $80.50 \pm \scriptstyle 0.26$ & $\textbf{75.31} \pm \scriptstyle 0.13$ & $\textbf{74.32} \pm \scriptstyle 0.52$ & $\textbf{73.27} \pm \scriptstyle 0.17$ \\
\addlinespace[3pt]
\midrule
 & \multicolumn{5}{c}{\textbf{Breast Cancer Wisconsin}} \\
\cmidrule(lr){2-6}
 & $p_{\text{miss}}=0$ & $p_{\text{miss}}=0.1$ & $p_{\text{miss}}=0.5$ & $p_{\text{miss}}=0.6$ & $p_{\text{miss}}=0.7$ \\
\midrule
Local   & \multicolumn{5}{c}{$86.36 \pm \scriptstyle 0.00$} \\
\midrule
SplitNN* & $93.37\pm \scriptstyle 0.00$ & $\textbf{93.37}\pm \scriptstyle 0.00$ & \cellcolor{red!15}$81.81\pm \scriptstyle 0.00$ & \cellcolor{red!15}$82.86\pm \scriptstyle 0.00$ & \cellcolor{red!15}$82.86\pm \scriptstyle 0.00$ \\
LASER*   & $86.69\pm \scriptstyle 0.00$ & \cellcolor{red!15}$84.90\pm \scriptstyle 0.00$ & \cellcolor{red!15}$85.06\pm \scriptstyle 0.00$ & \cellcolor{red!15}$84.74\pm \scriptstyle 0.00$ & \cellcolor{red!15}$86.19\pm \scriptstyle 0.00$ \\
\rowcolor{gray!20}MoPE    & $\textbf{94.15}\pm \scriptstyle 0.00$ & $93.30\pm \scriptstyle 0.00$ & $\textbf{90.37}\pm \scriptstyle 0.00$ & $\textbf{89.35}\pm \scriptstyle 0.00$ &  $\textbf{87.68}\pm \scriptstyle 0.00$\\ 
\bottomrule
\end{tabular}
\end{table}
\smallskip

\begin{figure}[h]
    \centering
    \begin{subfigure}{0.48\columnwidth}
        \centering
        \includegraphics[width=\textwidth]{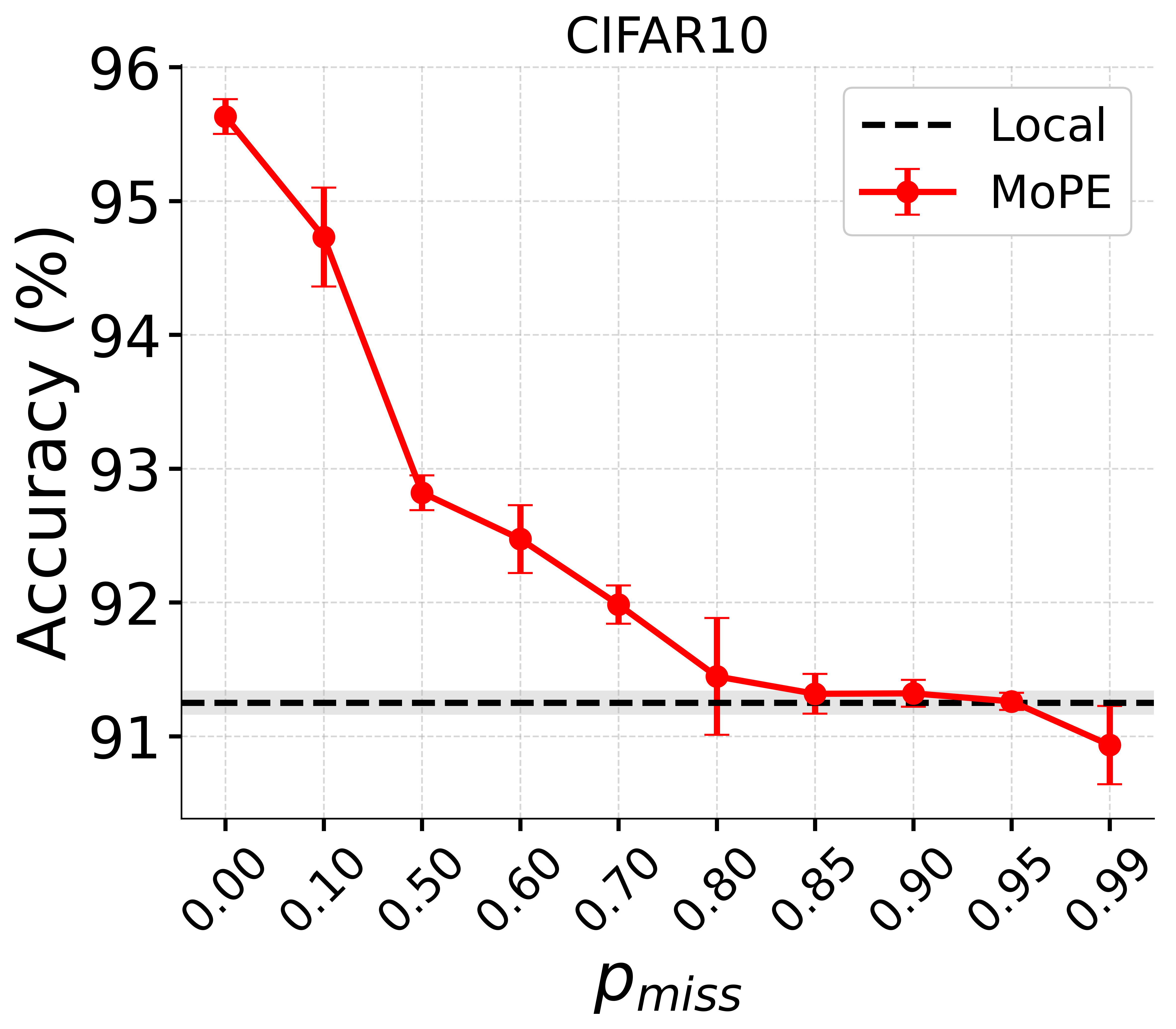}
        \caption{Accuracy increasing $p_{miss}$ in CIFAR10.}
        \label{subfig: extra missing cifar10}
    \end{subfigure}
    \hfill
    \begin{subfigure}{0.48\columnwidth}
        \centering
        \includegraphics[width=\textwidth]{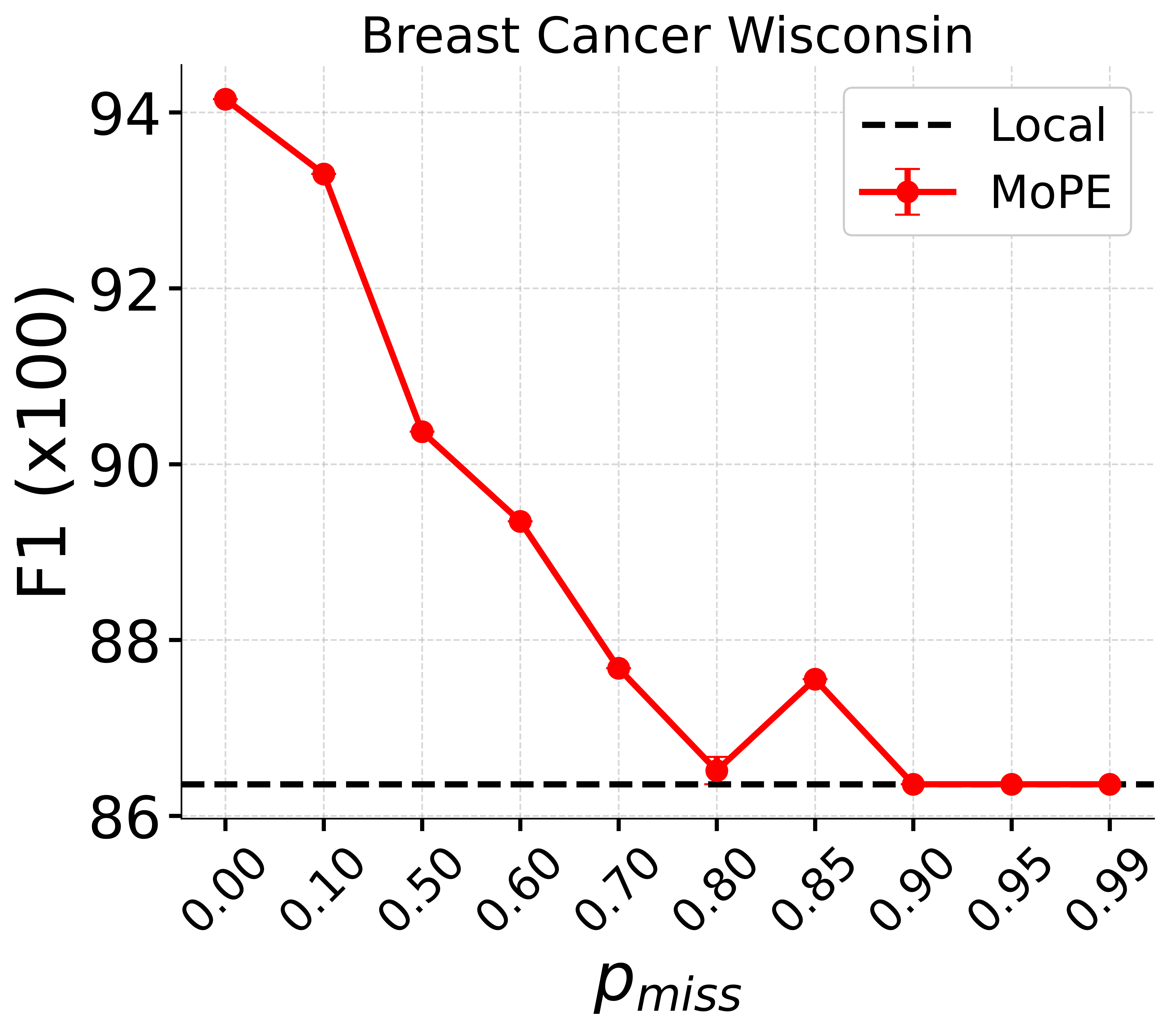}
        \caption{F1 increasing $p_{miss}$ in Breast Cancer Wisconsin.}
        \label{subfig: extra missing breast cancer}
    \end{subfigure}
 \caption{Evolution of MoPE's performance as $p_{miss}$ increases in CIFAR10 (a) and Breast Cancer Wisconsin (b).}
\label{fig: mope increased p miss}
\end{figure}

\noindent \textbf{Split-MoPE matches local classifier performance when no aligned data are available:} Table \ref{tab: results non noisy} also shows that Split-MoPE is the only VFL system whose performance does not degrade below the local baseline. To verify this, we extend the experiments to higher data-missingness rates in Figure \ref{fig: mope increased p miss}. The performance of the local classifier is represented by the horizontal dashed line. As shown, unlike other baselines, even in the most challenging scenarios ($p_{miss} = 0.99$), our approach never performs significantly worse than the local classifier, satisfying the requirements of an ideal VFL system (Section \ref{sec:method}). 
\smallskip

\noindent \textbf{Split-MoPE is robust against noisy or malicious parties:} To analyze the effect of noisy or malicious parties, we ran extensive experiments with CIFAR-10 and CIFAR-100, which are arguably the most challenging datasets, following \citep{khan2025vfl}. To simulate noisy or malicious parties, we introduce additional participants whose embeddings are sampled from a Normal distribution, $\mathcal{N}(0, 100)$. Note that by adding additional participants, while keeping the informative ones untouched, all the information remains on the federation. This allows us to isolate the effect of the noise. In particular, we set $p_{miss} = 0$ and $p_{miss} = 0.1$ for these experiments and test the performance of different VFL approaches in varying numbers of noisy or malicious parties. Table \ref{tab: performance delta noisy} shows the performance drop of the models, so the smaller the better. As can be seen, Split-MoPE shows less degradation across all cases, except for CIFAR-100 and 5 noisy participants, where SplitNN is slightly better. This can be due to the saturation in MoPE's routing mechanism. As the number of noisy participants increases relative to informative ones, the router faces greater difficulty distinguishing and routing input features to specialized experts. Surprisingly, LASER is outperformed by SplitNN under these conditions, which can be because of the aggregation mechanism. Specifically, our implementation of SplitNN uses concatenation, which preserves the spatial arrangement of each participant's contribution, potentially enabling the classification head to identify and reduce the influence of noisy inputs based on their fixed positions. In contrast, LASER relies on averaging, which blends embeddings together and disperses noise, likely preventing similar adaptive down-weighting.

\begin{table}[h]
\centering
\caption{Mean performance decay w.r.t. scenario without noisy participants on CIFAR-10 and CIFAR-100 datasets. Our approach is shaded in {\setlength{\fboxsep}{0pt}\colorbox{gray!20}{gray}} and smallest differences are highlighted in \textbf{bold}. MoPE consistently minimizes performance loss across all scenarios, demonstrating superior robustness to noisy participants compared to baselines.}
\small
\label{tab: performance delta noisy}
\begin{tabular}{lrrrrr}
\toprule
 & \multicolumn{5}{c}{\textbf{CIFAR-10, $p_{\text{miss}}=0$}} \\
\cmidrule(lr){2-6}
 & 1 noisy & 2 noisy & 3 noisy & 4 noisy & 5 noisy \\
\midrule
SplitNN* & 3.05 & 4.67 & 5.28 & 5.47 & 5.92 \\
LASER*   & 3.76 & 6.46 & 11.45 & 16.30 & 22.31 \\
\rowcolor{gray!20}MoPE    & \textbf{0.35} & \textbf{0.18} & \textbf{2.75} & \textbf{3.43} & \textbf{3.53} \\
\addlinespace[3pt]
\midrule
 & \multicolumn{5}{c}{\textbf{CIFAR-10, $p_{\text{miss}}=0.1$}} \\
\cmidrule(lr){2-6}
 & 1 noisy & 2 noisy & 3 noisy & 4 noisy & 5 noisy \\
\midrule
SplitNN* & 3.00 & 4.18 & 4.80 & 19.10 & 19.26 \\
LASER*   & 4.44 & 7.32 & 12.47 & 18.32 & 22.31 \\
\rowcolor{gray!20}MoPE    & \textbf{1.65} & \textbf{2.57} & \textbf{3.08} & \textbf{4.35} & \textbf{3.95} \\
\addlinespace[3pt]
\midrule
 & \multicolumn{5}{c}{\textbf{CIFAR-100, $p_{\text{miss}}=0$}} \\
\cmidrule(lr){2-6}
 & 1 noisy & 2 noisy & 3 noisy & 4 noisy & 5 noisy \\
\midrule
SplitNN* & 8.34 & 12.99 & 15.43 & 16.94 & \textbf{18.81} \\
LASER*   & 14.61 & 24.62 & 31.77 & 36.34 & 40.18 \\
\rowcolor{gray!20}MoPE    & \textbf{3.15} & \textbf{4.92} & \textbf{4.98} & \textbf{10.30} & 20.38 \\
\addlinespace[3pt]
\midrule
 & \multicolumn{5}{c}{\textbf{CIFAR-100, $p_{\text{miss}}=0.1$}} \\
\cmidrule(lr){2-6}
 & 1 noisy & 2 noisy & 3 noisy & 4 noisy & 5 noisy \\
\midrule
SplitNN* & 8.48 & 12.56 & 15.59 & 16.50 & 18.14 \\
LASER*   & 14.05 & 23.44 & 29.98 & 35.39 & 38.57 \\
\rowcolor{gray!20}MoPE    & \textbf{3.36} & \textbf{5.69} & \textbf{5.44} & \textbf{11.04} & \textbf{15.41} \\
\bottomrule
\end{tabular}
\end{table}

To support this hypothesis, additional experiments were conducted on CIFAR-10 using SplitNN with averaging instead of concatenation. The accuracy dropped from  $95.27\pm \scriptstyle 0.04$ to  $14.84\pm \scriptstyle 0.69$ upon introducing a single noisy participant. This highlights the importance of aggregation and shows that LASER's sampling mechanism mitigates the negative impact of noise. 


\section{Analysis of the results}
\label{sec:analysis}
In this section, we analyze deeper the behavior of Split-MoPE, focusing on understanding the weights assigned by the router to different experts with noisy or malicious parties, as well as analyzing the effect of pretrained encoders in terms of communication efficiency and performance.
\smallskip

\begin{table*}[]
\centering
\caption{Mean weight given by the router to each expert on different scenarios on CIFAR-10 with full sample alignment. $A$ is the active participant, $P$ the passive one and $N_n$, the $n^{th}$ noisy participant. \ding{55} indicates that the expert does not exist.}
\label{tab: router weights}
\begin{tabular}{lccccccc}
\toprule
\textbf{Scenario} & \multicolumn{7}{c}{\textbf{Expert alignment}} \\
\cmidrule(lr){2-8}
 & $\{A\}$ & $\{P, A\}$ & $\{N_1, A\}$ & $\{P, N_1, A\}$ & $\{N_2, A\}$ & $\{N_1, N_2, A\}$ & $\{P, N_1, N_2, A\}$ \\
\midrule
No Noisy & 0 & 0.99 & \ding{55} & \ding{55} & \ding{55} & \ding{55} & \ding{55} \\
1 Noisy  & 1.00 & 1.00 & 0 & 0 & \ding{55} & \ding{55} & \ding{55} \\
2 Noisy  & 1.00 & 1.00 & 0 & 0 & 0 & 0 & 0 \\
\bottomrule
\end{tabular}
\end{table*}

\noindent \textbf{Split-MoPE reliably identifies noisy or malicious parties as shown by the weights assigned by the router:} To better understand the robustness of Split-MoPE we analyzed the routing weights assigned by the router for CIFAR-10 with $p_{miss} = 0$ and represent them in Table \ref{tab: router weights}. The results show that the router effectively identifies noisy participants and prevents experts that process their information from contributing to the final prediction. This behavior explains the model’s robustness under noisy conditions. Moreover, because the MoPE head is interpretable, we can identify why performance decays when noisy participants are introduced, even though we have shown that the router identifies and ignores them: it equally weights both informative experts, which is not optimal. This suggests that noisy participants do actually have a negative effect on the training process of Split-MoPE.


Analyzing the weights assigned to each expert reveals the relative influence of each collaborator on the model's prediction. Although the exact reasoning behind a prediction remains unknown, it is possible to identify which participant or group contributed to the decision. Defining $w_e$ as the weight given to the $e^{th}$ expert that processes the embeddings of a specific subset of $\tilde{\mathcal{P}}(\mathcal{K})$, the contribution, $C$, of the $k^{th}$ participant is computed via Equation \ref{equ: party contribution}:


\begin{equation}
    \label{equ: party contribution}
    C_k=\sum_{e\in\tilde{\mathcal{P}}(\mathcal{K})}\frac{\mathds{1}_{k\in e}w_e}{w_e}
\end{equation}

For scenarios where the active participant wishes to measure contributions exclusively from the passive participants, Equation \ref{equ: party contribution} can be adapted to exclude the expert processing embeddings solely from the active participant, i.e., summing over $\tilde{\mathcal{P}}(\mathcal{K})\backslash\{K\}$.

These metrics are valuable not only for interpreting the provenance of a prediction but also for establishing equitable incentive mechanisms. By quantifying the relative value of each passive participant's data, Split-MoPE provides a framework for fair compensation within the collaborative ecosystem.


Note that while Table \ref{tab: router weights} reports the mean weights, the same exact analysis can be done per-sample, thus making it possible to analyze who has contributed to each individual prediction.
\smallskip

\noindent \textbf{Pretrained encoders enable drastically reducing the communication footprint of VFL training:} On split learning schemes, local training or pretrained models are rarely employed. As earlier stated in Section \ref{sec: background}, multiple communication rounds are required to train the models.

The total communication overhead incurred during the training process is the sum of the costs of performing the required forward passes and backpropagations, which grows linearly with both the dataset size and the number of epochs. Equation \ref{equ: split training} represents the total cost of performing the needed forward and backward communications, assuming a precision of 4 bytes:

\begin{equation}
    \label{equ: split training}
    \text{comm\_cost} = 2\cdot(K-1)\cdot\text{epoch}\cdot|\mathcal{D}|\cdot z \cdot4\;\text{(bytes)}
\end{equation}

where,
\begin{description}
    \item $K$ denotes the number of participants.
    \item[] $|\mathcal{D}|$ denotes the aligned dataset's size.
    \item $z$ denotes the embedding size.
\end{description}

Alternatively, pretrained backbones require only a single forward pass per training, making overall efficiency $2 \times $\textit{epochs} greater. 

To illustrate this drastic reduction, consider a collaboration of two entities with $25,000$ aligned samples, training for $100$ epochs using an embedding size of $384$. The total communication footprint exceeds $7.68$ GB. Alternatively, decoupled training results in a process that exchanges $38.4$ MB of data, yielding a $200\times$ reduction.
\smallskip

\noindent \textbf{More capable pretrained encoders bring extra performance and robustness to non-aligned data:} To understand how pretrained encoders affect the performance of Split-MoPE, we select three encoder sizes for DINO-v2 and Qwen3 and test Split-MoPE for CIFAR-10 and Breast Cancer Wisconsin under varying values of $p_{miss}$. The results in Table \ref{tab: results backbones} show that, as expected, larger encoders perform consistently better for both datasets. But somehow surprisingly, larger encoders also show higher robustness to non-aligned data. For CIFAR-10, ViT-S (21M parameters) loses almost 4 absolute points, whereas ViT-L (300M parameters) loses only 2 points. A more pronounced difference is observed for Breast Cancer Wisconsin, where the smallest encoder drops by 11 points, while the largest one only drops by 3 points.

\begin{table}[h]
\centering
\caption{Test accuracy (\%) on CIFAR-10 and F1 score ($\times100$) on Breast Cancer Wisconsin under different missing data probabilities $p_{\text{miss}}$ with different backbones. Mean $\pm$ std. dev. over three runs is reported.}
\small
\label{tab: results backbones}
\begin{tabular}{lccc} 
\toprule
 & \multicolumn{3}{c}{\textbf{CIFAR-10}} \\ 
\midrule 
 & $p_{\text{miss}}=0$ & $p_{\text{miss}}=0.1$ & $p_{\text{miss}}=0.7$ \\
\midrule
ViT-S & $95.63 \pm \scriptstyle 0.13$ & $94.73 \pm \scriptstyle 0.37$ & $91.98 \pm \scriptstyle 0.14$ \\
ViT-B & $97.05 \pm \scriptstyle 0.07$ & $96.62 \pm \scriptstyle 0.14$ & $93.96 \pm \scriptstyle 0.05$ \\
ViT-L & $98.45 \pm \scriptstyle 0.04$ & $96.90 \pm \scriptstyle 1.81$ & $96.38 \pm \scriptstyle 0.12$ \\
\addlinespace[3pt]
\midrule
 & \multicolumn{3}{c}{\textbf{Breast Cancer Wisconsin}} \\ 
\midrule 
 & $p_{\text{miss}}=0$ & $p_{\text{miss}}=0.1$ & $p_{\text{miss}}=0.7$ \\
\midrule
Qwen3-0.6B & $94.22 \pm \scriptstyle 0.0$ & $92.18 \pm \scriptstyle 0.44$ & $83.05 \pm \scriptstyle 0.43$ \\
Qwen3-4B   & $94.15 \pm \scriptstyle 0.0$ & $93.30 \pm \scriptstyle 0.0$ & $87.68 \pm \scriptstyle 0.0$ \\
Qwen3-8B   & $94.22 \pm \scriptstyle 0.0$ & $93.37 \pm \scriptstyle 0.0$ & $91.55 \pm \scriptstyle 0.11$ \\
\bottomrule
\end{tabular}
\end{table}

\section{Conclusions and Future Work}


In this work, we introduce Split-MoPE, a novel Vertical Federated Learning (VFL) approach designed for realistic scenarios with partial sample alignment among collaborators. Our extensive evaluation demonstrates that Split-MoPE consistently outperforms state-of-the-art baselines LASER and SplitNN, particularly as data alignment is reduced (e.g. improvements of over 1\% in accuracy and more than 5 points in F1 score) and under noisy/malicious participants (e.g., up to $25\times$ less degradation).

The proposed Mixture of Predefined Experts (MoPE) not only boosts robustness but also enables per-prediction contribution quantification, thereby enhancing explainability and fair reward mechanisms in federations. Designed to require only a single communication round, Split-MoPE substantially reduces the overall footprint compared to multi-round end-to-end training schemes.

For future work, we plan to explore alternative training strategies to further enhance the performance of the MoPE. Given the architecture's modularity, we also aim to investigate Split-MoPE in heterogeneous multimodal environments, assessing how the weighting mechanism prioritizes diverse data sources (e.g., vision vs. tabular) during ensemble predictions.

\begin{acks}
This research was supported by the Ministry of Science and Innovation of the Spanish Government (DeepKnowledge project PID2021-127777OB-C21), the Basque Government (IXA excellence research group IT1570-22), the CHIST-ERA grant (Project Geo-R2LLM, CHIST-ERA-23-MultiGIS-04) funded by the the Ministry of Science and Innovation of the Spanish Government (PCI2025-163286) and the European Union under Horizon Europe (Project LUMINOUS, grant number 101135724 and Project FULL-MAP, grant number 101192848).
\end{acks}

\bibliographystyle{ACM-Reference-Format}
\bibliography{sample-base}

\appendix

\section{Models, Hyperparameters and Datasets}\label{appendix: arch and params}

\noindent \textbf{Architectural details:}
In Table \ref{tab: backbone params}, we summarize the main characteristics of the backbones that have been used in the experiments of the paper.

\begin{table}[h]
\centering
\caption{Parameter count, \#Params, as well as embedding dimensionality, $\mathbf{z}_{dim}$, of the employed pretrained backbones for vision and tabular datasets. The ones used in the main experiments are shaded in {\setlength{\fboxsep}{0pt}\colorbox{gray!20}{gray}}.}
\small
\label{tab: backbone params}
\begin{tabular}{lccc} 
\toprule
 & \multicolumn{2}{c}{\textbf{Vision}} \\ 
\midrule 
 & \#Params & $\mathbf{z}_{dim}$\\
\midrule
\rowcolor{gray!20}ViT-S & $21M$ & $384$ \\
ViT-B & $86M$ & $768$ \\
ViT-L & $300M$ & $1024$ \\
\addlinespace[3pt]
\midrule
 & \multicolumn{2}{c}{\textbf{Tabular}} \\ 
\midrule 
 & \#Params & $\mathbf{z}_{dim}$ \\
\midrule
Qwen3-0.6B & $0.6B$ & $1024$ \\
\rowcolor{gray!20}Qwen3-4B   & $4B$ & $2560$ \\
Qwen3-8B   & $8B$ & $4096$ \\
\bottomrule
\end{tabular}
\end{table}

As classifiers, we have employed two-layer NNs, that have $\mathbf{z}_{dim}$ as input, double its dimension, i.e., $2 \times \mathbf{z}_{dim}$ with ReLU activation and a final layer whose size is defined by the number of classes.
\smallskip

\noindent \textbf{Hyperparameters:} Every model has been trained using Adam \citep{adam2014method} as optimizer. Every hyperparameter has been set to its default value, but the learning rate is set to $0.0001$. For the vision datasets, the selected batch size was 256, whereas in Breast Cancer Wisconsin, batches of 25 samples were employed.
\smallskip

\noindent \textbf{Datasets:} For CIFAR-10 and CIFAR-100 we have followed LASER \citep{valdeira2024vertical}, without modifying its original code. For Breast Cancer Wisconsin, $80\%$ of the samples have been used for training, whereas the remaining $20\%$ has been reserved for testing.

\section{Additional Results Comparing End-to-End Training and Frozen Backbones} \label{appendix: cifar10 frozen}
In Table \ref{tab: end to end vs frozen appendix}, we report the results using trainable and frozen (marked with an asterisk) backbones on CIFAR-10.

\begin{table}[h]
\centering
\caption{Test accuracy (\%) CIFAR-10 performing end-to-end training and employing frozen backbones (*). Mean $\pm$ std. dev. over three runs is reported. Best results in bold.}
\small
\label{tab: end to end vs frozen appendix}
\begin{tabular}{lccc} 
\toprule
 & \multicolumn{3}{c}{\textbf{CIFAR-10}} \\ 
\midrule 
 & $p_{\text{miss}}=0$ & $p_{\text{miss}}=0.1$ & $p_{\text{miss}}=0.7$ \\
\midrule 
LASER & $88.35 \pm \scriptstyle 2.19$ & $86.31 \pm \scriptstyle 2.32$ & $73.78 \pm \scriptstyle 1.63$ \\
LASER* & $\textbf{95.42} \pm \scriptstyle 0.02$ & $\textbf{94.66} \pm \scriptstyle 0.42$ & $\textbf{91.91} \pm \scriptstyle 0.29$ \\
\midrule
SplitNN & $80.61 \pm \scriptstyle 1.35$ & $79.40 \pm \scriptstyle 0.90$ & $80.37 \pm \scriptstyle 1.41$ \\
SplitNN* & $\textbf{95.61} \pm \scriptstyle 0.10$ & $\textbf{95.55} \pm \scriptstyle 0.08$ & $\textbf{90.59} \pm \scriptstyle 0.22$ \\
\bottomrule
\end{tabular}
\end{table}

These results are consistent with the ones reported in the main paper (see Table \ref{tab: end to end vs frozen}). They show that end-to-end training with pretrained backbones is detrimental. It results in a drop of over 18\% in accuracy in the most challenging scenarios.

\section{Improved Results w.r.t. LASER} \label{appendix: laser improvement}
As stated in Section \ref{sec:results}, our modified version of LASER \citep{valdeira2024vertical} has yielded significant performance improvements w.r.t what was reported on the original paper. These differences have been presented in Table \ref{tab: deltas laser}.

\begin{table}[]
\centering
\caption{Test accuracy (\%) on CIFAR-10 and CIFAR-100. Results reported in this paper (*) and the ones reported in the original paper are compared. $\Delta\uparrow$ is the mean improvement of our results.}
\small
\label{tab: deltas laser}
\begin{tabular}{lccc} 
\toprule
 & \multicolumn{3}{c}{\textbf{CIFAR-10}} \\ 
\midrule 
 & $p_{\text{miss}}=0$ & $p_{\text{miss}}=0.1$ & $p_{\text{miss}}=0.5$ \\
\midrule
LASER & $91.50 \pm \scriptstyle 0.10$ & $90.20 \scriptstyle \pm 0.40$ & $79.40 \pm \scriptstyle 1.60$ \\
LASER* & $95.42 \pm \scriptstyle 0.02$ & $94.73 \pm \scriptstyle 0.37$ & $92.82 \pm \scriptstyle 0.13$ \\
$\Delta\uparrow$ & $3.92$ & $4.53$ & $13.12$ \\
\addlinespace[3pt]
\midrule
 & \multicolumn{3}{c}{\textbf{CIFAR-100}} \\ 
\midrule 
 & $p_{\text{miss}}=0$ & $p_{\text{miss}}=0.1$ & $p_{\text{miss}}=0.5$ \\
\midrule
LASER & $72.30\pm \scriptstyle 0.10$ & $68.90 \pm \scriptstyle 0.60$ & $51.90 \pm \scriptstyle 1.20$ \\
LASER*   & $82.26\pm \scriptstyle 0.06$ & $79.83 \pm \scriptstyle 0.50$ & $74.52 \pm \scriptstyle 0.54$ \\
$\Delta\uparrow$ & $9.96$ & $10.93$ & $22.62$ \\
\bottomrule
\end{tabular}
\end{table}

Note that the results are not directly comparable since the backbones are not identical. In the original implementation, the authors employed a ResNet18 \citep{he2016deep} as a feature extractor, which has a total of 11.7M parameters. On the other hand, our ViT-s has 21M parameters.

Nevertheless, as shown in Table \ref{tab: deltas laser}, the results we have used as a baseline are up to 22\% better, representing a non-negligible improvement. Furthermore, this supports the finding of Section \ref{sec:analysis}: larger local encoders are more robust to data missingness.

\section{Performance Plots With Malicious Participants} \label{appendix: performance}
\begin{figure}[]
    
    \begin{subfigure}{\columnwidth}
        \centering
        \includegraphics[width=\columnwidth]{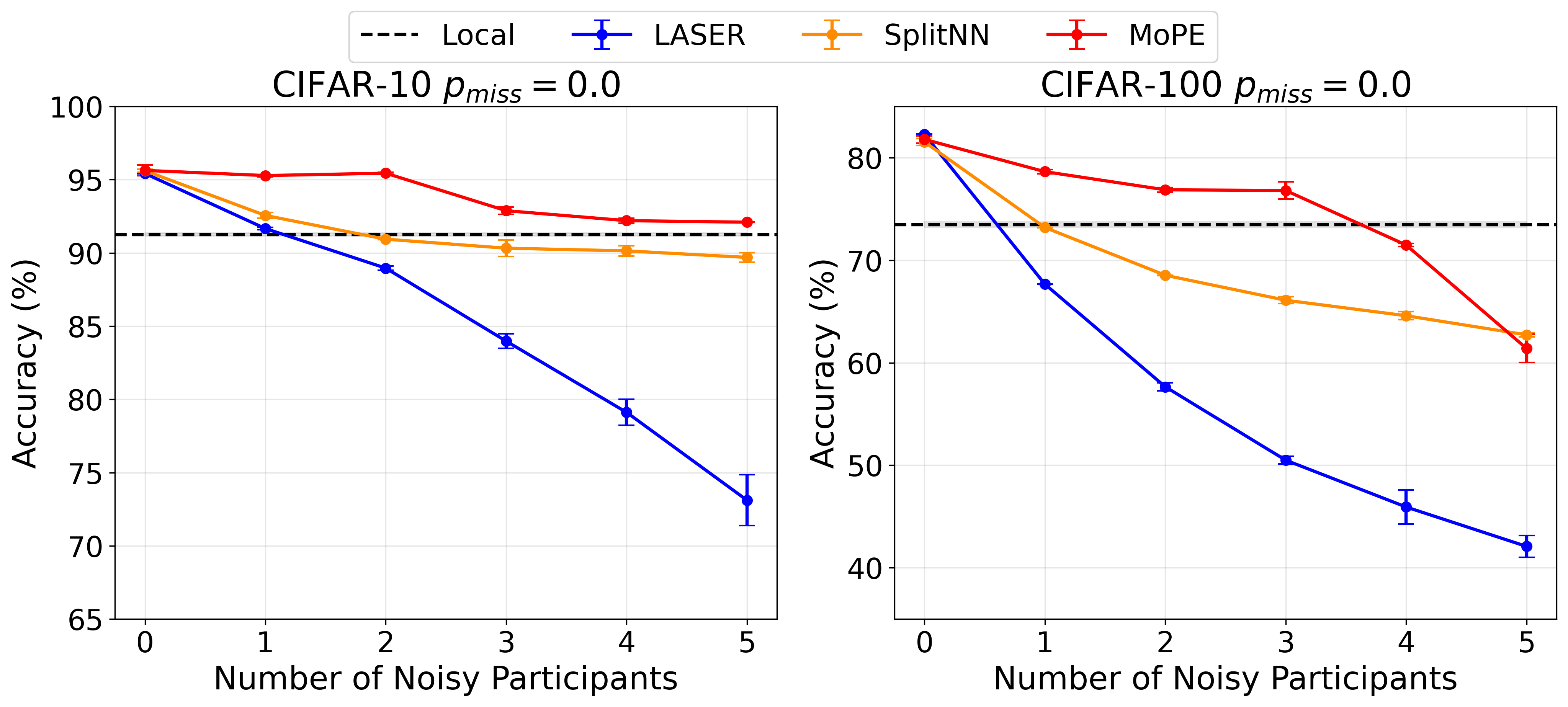}
        \caption{Results with multiple noisy participants with $p_{miss}=0.0$.}
        \label{subfig: performance noisy 00}
    \end{subfigure}
    
    \vspace{0.3cm}
    
    \begin{subfigure}{\columnwidth}
        \centering
        \includegraphics[width=\columnwidth]{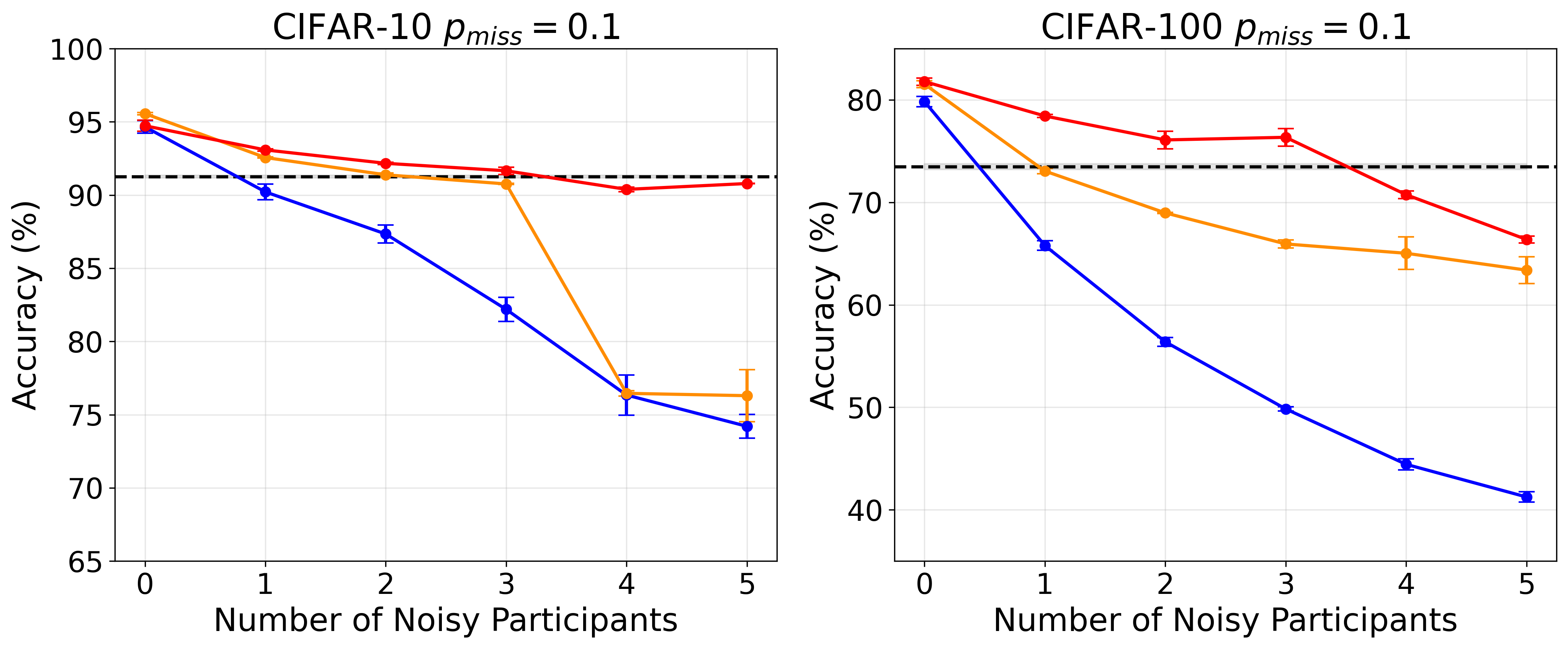}
        \caption{Results with multiple noisy participants with $p_{miss}=0.1$.}
        \label{subfig: performance noisy 01}
    \end{subfigure}
    
    \caption{Results with multiple noisy participants varying $p_{miss}$ on CIFAR-10 and CIFAR-100.}
    \label{fig: noisy participants}
\end{figure}

In Figure \ref{fig: noisy participants}, we show the performance of each approach instead of the deltas reported in Table \ref{tab: performance delta noisy}. Note that the values reported in Table \ref{tab: performance delta noisy} are the result of subtracting the first value of each plot, i.e., the value when the number of noisy participants is 0, from all other results.

It is worth noting that across all tested scenarios, our approach, MoPE, consistently improves the model trained locally in settings with up to 3 noisy participants. Given that there are two informative participants, our approach outperforms locally trained models in scenarios where more than half of the participants in the federation are non-informative or malicious.

\section{On the Usage of Text Embedding Models to Process Tabular Data}\label{appendix: tabular dataset}

\noindent \textbf{Methodological overview:} While text embedding models are traditionally employed for natural language processing, they can process tabular data when it is properly serialized as text. Our suggested approach transforms a structured data row into a string before passing it through a pre-trained transformer-based encoder.

\begin{figure}[h]
  \centering
  \includegraphics[width=\linewidth]{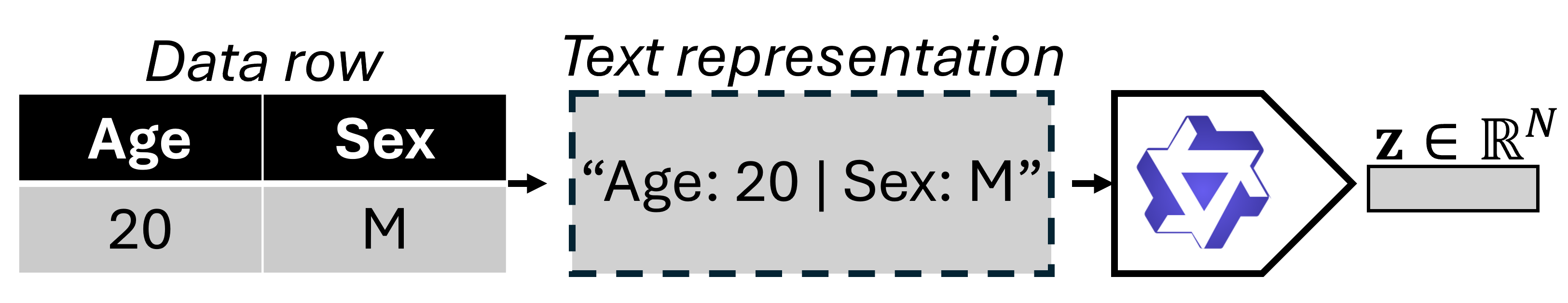}
  \caption{Visual representation of the suggested process to generate embeddings of tabular datasets using text embedding models. The data row is converted into text and processed by a text embedding model.}
  \label{fig: tabular embedding}
\end{figure}

As illustrated in Figure \ref{fig: tabular embedding}, each sample, row, is treated as a sentence. This allows the model to process structured information as if it were regular text, producing an $N$-dimensional embedding.

\noindent \textbf{Empirical validation:} Because processing tabular datasets as text is not a common approach, we benchmarked its performance using the Breast Cancer Wisconsin dataset. We utilized a Logistic Regression classifier as a linear probe to evaluate whether the resulting embeddings retain the \textit{informativeness} of raw data.

\begin{table}[h]
\centering
\caption{Test F1 score ($\times 100$) on the Breast Cancer Wisconsin dataset. Results compare the use of all original features with the proposed Tabular Embeddings using Logistic Regression as the classifier.}
\small
\label{tab: breast cancer f1 tab embedding}
\begin{tabular}{lc} 
\toprule
\textbf{Feature Type} & \textbf{Test F1 ($\times 100$)} \\ 
\midrule
Original & $98.61$ \\
Embeddings & $93.71$ \\
\bottomrule
\end{tabular}
\end{table}

The results in Table \ref{tab: breast cancer f1 tab embedding} demonstrate that the embeddings are highly \textit{informative}, maintaining an F1 score above 93\%. This shows that processing tabular datasets as text is feasible for the VFL context, where raw data cannot be exchanged.

\end{document}